# Recognition Of Surface Defects On Steel Sheet Using Transfer Learning

Jingwen Fu , Xiaoyan Zhu* , and Yingbin Li

Xian Jiaotong University, Shaanxi , China
*fu1371252069@stu.xjtu.edu.cn,xjtueden@163.com,851576881@qq.com*

## Abstract

*Automatic defect recognition is one of the research hotspots in steel production, but most of the current methods mainly extract features manually and use machine learning classifiers to recognize defects, which cannot tackle the situation, where there are few data available to train and confine to a certain scene. Therefore, in this paper, a new approach is proposed which consists of part of pretrained VGG16 as a feature extractor and a new CNN neural network as a classifier to recognize the defect of steel strip surface based on the feature maps created by the feature extractor. Our method achieves an accuracy of 99.1% and 96.0% while the dataset contains 150 images each class and 10 images each class respectively, which is much better than previous methods.*

## 1. Introduction

Undoubtedly, steel strip is one of the most widely used materials in modern[22][4][20][13][9] industry. However, steel strip may be affected by materials, equipment, etc. during the production process, resulting in various defects[15][35] on its surface, e.g., scratches, surface crazing and rolled-in scale, which can cause huge economic losses to the enterprise. Therefore, real-time accurate inspection of surface defects has become an indispensable section in the iron and steel enterprises.

In the early 1990s, the traditional methods of surface inspection mainly include artificial visual inspection, magnetic flux leakage testing[19], etc. Due to the influence of the subjective factor and high error inspection rate, these methods cannot meet the requirement of real-time[17]. Recently, the visual-based inspection technology, as a kind of non-contact inspection method, has become a research hotspot in the field of surface defects inspection. This method integrates many advanced technologies including image processing, optics, pattern recognition, artificial intelligence, hence it has obtained real-time, high accuracy rate and reliability. Given the characteristics of the real-time and easy to realize intelligently, this technology has been widely used in the online real-time inspection of the automatic production lines such as paper[1], glass[3], and steel bar[16].

The traditional automated surface inspection system of the steel consists of four subsystems: image acquisition, rapid detection, feature extraction, and defect classification, which is showed clearly in figure 1. As we can see from Figure 1, image acquisition is to make the defects visible. Rapid detection is to acquire defect images and locate the positions of the defects. Then the feature extraction, as one of the important tasks of a surface inspection system, is mainly to describe the defect characteristics. The last component is always called defect classification. In this subsystem, the extracted features are attached to class labels employing a classifier. The first two components have been studied very well[25][21][34][33][2]. Therefore, the main task of the steel surface inspection system in the current period is to improve the defect recognition rate, which includes feature extraction and defect classification.

In the past, three methods named the LBP[23] and the completed local binary pattern (CLBP)[8] and local ternary patterns (LTP)[30] have achieved great success in the feature extraction, however, their threshold schemes are sensitive to noise. In 2013, K.song proposed a noise-robust method named AECLBP[28], in which an adjacent evaluation window is constructed to modify the threshold scheme of the completed local binary pattern (CLBP), to improve the recognition rate and get the more robust feature descriptor against noise. Unfortunately, this method can also only achieve moderate recognition accuracy in the toughest situation with additive Gaussian noise. What′s worse, AE-

---
*Corresponding author.

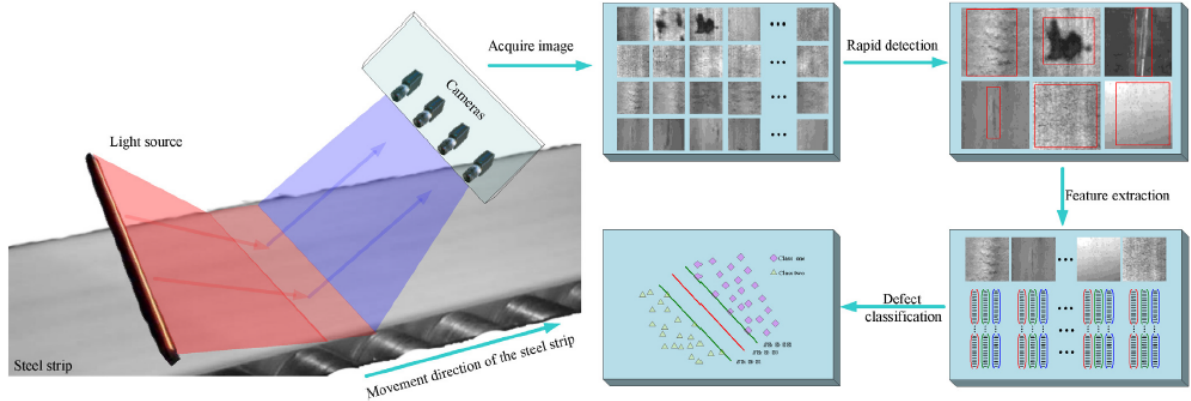

Figure 1. Traditional structure of automated surface inspection system.

CLBP needs to set some professional parameters such as the size of the evaluation window, which is very inconvenient for users.

Fortunately, in recent years, deep learning technology in the field of pattern recognition emerged and has already been widely used. Deep learning technology greatly improves the accuracy of image classification. Among these neural networks, the more famous ones are AlexNet[14], VGGNet[27], googleNet[29] and resNet[10], which can help us to do the recognition of steel surface defects.

However, powerful deep learning techniques have not been widely used in the steel strip surface defect recognition. The most important constraint is that high-quality images with labels that can be used for the training model are often not enough because they are difficult to obtain, especially in the inspection of surface defects.

Interestingly, Yosinski et al. published an article named "How transferable features in deep neural networks[32] in 2014, which concludes that the shallow layer of the neural network mainly extracts simple common features of the image, such as edge information, and the parameters of these layers have little difference in the values of different data sets, which we can use to build our image extraction of the surface inspection system.

Inspired by these papers, we propose a steel strip surface defect recognition method based on transfer learning. Our algorithm first extracts the features of the steel strip surface image using the image feature extractor VGG16 that has already trained the parameters using hundreds of thousands of pictures from imagenet and then accesses a convolutional neural network to finally train a classifier.

In order to make our experiment more convincing, we also compared our method with two other methods based on deep learning named CSDS[35] and PLCNN[5], which were respectively proposed in 2017 and 2020. We did many experiments and compared the results with previous studies. The data set taken in this experiment is the NEU surface defect database collected by the teachers of the Northeastern University of China.

Experiments show that our accuracy reaches 0.99 in this dataset, more meaningful, the algorithm proposed by us, which is built on a very small training dataset with only 10 training instances each class, reached 0.90 of the classification accuracy, whose test data has 150 instances each class, while the past studies can only reach approximately 0.75. And obviously, our method is meaningful to solve the problem of scarcity of data sets in other industrial fields, not only for the inspection of steel surface defects.

The remaining part of the paper is organized as follows sections: Section 2 provides a brief review of the related work. Section 3 shows the inspection method in detail. Section 4 provides the experimental process and the corresponding results. The conclusion of our work is given in Section 5.

## 2. Related Work

### 2.1. Steel Strip Surface Defect Detection

In the early 1990s, the traditional methods of surface inspection mainly include artificial visual inspection, magnetic flux leakage testing[19], etc. Due to the influence of the subjective factor and high error inspection rate, these methods cannot meet the requirement of real-time[17].

In 2013, K. Song et al. used the AECLBP-based feature extraction method[28] and the SVM classifier to classify steel strip defects, which is shown in the following figure 1, and achieve an accuracy of 0.9893. However, the method uses the knowledge of a specific field to extract features, which makes the method difficult for other steel products or other steel defects, and even other fields.

In 2016, Lei Wang put forward a new way[31] to detect the defect in steel strip. It works well, however, 3D information is needed. It will need extra time and resources to obtain 3D information.

In 2017, S Zhou et al, proposed a classification[35] based on the convolutional neural network, which performed very good. However, it needs lots of training data sets and the author didn't analyze the effects of noise and the size of training data sets.

In 2019, Zheng Liu et al, put forward a deep neural network Inception Dual Network for steel strip defect detection[18], which can achieve high accuracy on the steel strip defect dataset. But it still needs a lot of data to train the neural network. When the dataset is limited to a relatively small size, it can't show a very good performance.

## 2.2. Transfer Learning

Sinno Jialin Pan and others published a review of transfer learning In 2009[24]. It shows that transfer learning technology can fully utilize the knowledge that the neural network learns on other data sets, so it can greatly reduce the demand of neural networks for data sets to train. Another scholar named Hoo-Chang Shin, successfully demonstrated in his research that the convolutional neural network uses the powerful cross-domain application ability to transfer learning in 2016.

In 2017, Kasthurirangan et al. used the first 15 layers of vgg16 as feature extractors[7], and replaced the last few fully connected layers of vgg16 with traditional machine learning methods such as support vector machines, random forests, and multilayer perceptrons. A small data set of 760 images of road defects achieved 0.90 accuracy. However, in the paper, the principle of the neural network is not analyzed in detail, and the method of data enhancement is not used, which is important to solve the practical problems.

In 2018, Luke Scime et al.[26] used pre-trained AlexNet to keep all parameters fixed and retrained the final fully connected layer to achieve detection and classification of laser powder fusion additives. In order to adapt the grayscale image to the shape of the RGB image input by AlexNet, the author does not simply copy the grayscale image three times, but adds different scale features in different channels.

Based on the research in recent years, we can find that the use of transfer learning with deep learning has become a research hotspot. Using a large amount of data in a certain field to train the image feature extraction model, and then applying this model to the identification of steel strip surface defects, can well solve the shortcomings of the number of industrial steel strip surface images. This study used the VGG16 model built by imagenet data set for image feature extraction and connected to a convolutional neural network to construct a classification model.

## 3. Surface Defects Recognition System

### 3.1. Overview

The traditional structure of the automated surface inspection system has been shown in Figure 1. The system mainly consists of four subsystems: image acquisition, rapid detection, feature extraction, and defect classification.

Like tradition methods, the approach proposed by us to recognize the surface defect of steel also contains four stages. Due to the good result of image acquisition, rapid detection in previous methods, here we just use the same method in image acquisition, rapid detection. Figure 3 shows the overall structure of our neural network which consists of two parts: feature extractor and classifier. The extractor uses the part of pretrained vgg16 and its weights don't update during the training process. Its function is to extract the features of the picture and transform them into feature maps. The second part is a classifier, whose function is to class this defect based on the feature maps created by the extractor. In the following, this paper will describe this structure in detail.

### 3.2. Image acquisition and rapid detection subsystem

The hardware configuration of the image acquisition mainly includes two components: light source and cameras. The light source provides illumination to make the defects visible. Since the light-emitting diode (LED) has many advantages such as little power and longevity, it is used to produce the narrow-spectrum light in this work. In order to cover the width of the steel strip, four area scan CCD cameras are used. The size of the original grayscale images is $1024 * 1024$ pixels. In this work, in order to save calculation time for subsequent sections, the original images are performed on the downsampling process, i.e., the sampled images are set as $200 * 200$ pixels. And many rapid detection methods available, like [25] [21], which all show

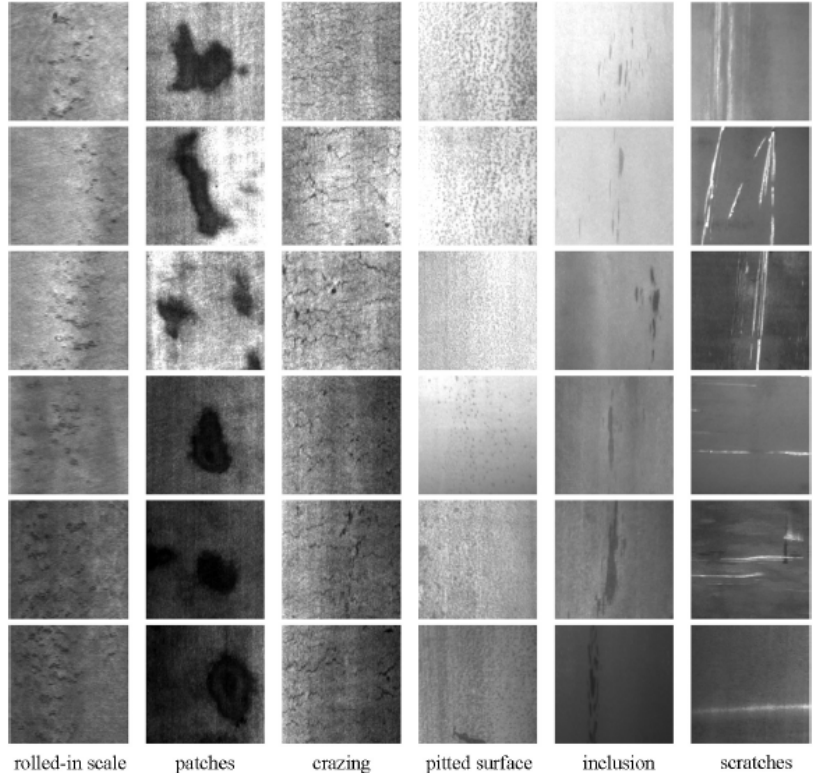

Figure 2. Samples of six kinds of typical surface defects on NEU surface defect database. Each row shows one example image from each of 300 samples of a class

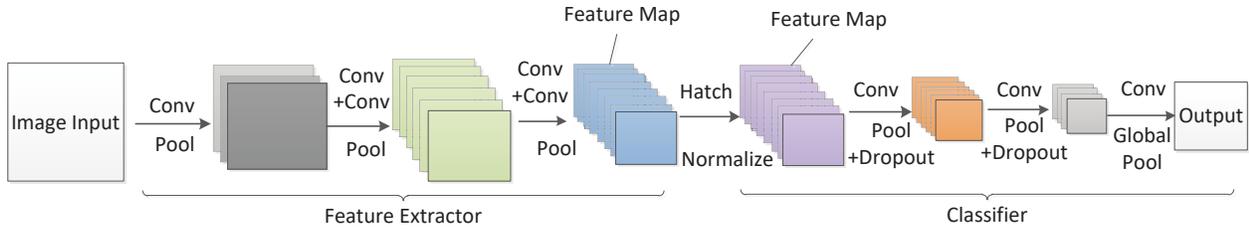

Figure 3. The structure of the SSDR neural network

### 3.3. Feature Extractor subsystem

VGGNet[27] is proposed by K. Simonyan, which is an improved version of AlexNet. VGGNet has two versions: VGG16 (with 16 parameter layers) and VGG19. VGGNet uses three 3x3 convolution cores instead of 7x7 convolution kernels, and two 3x3 convolution cores instead of 5 x 5 convolution kernels, which improves the depth of the network and, to some extent, the effectiveness of the neural network, with the same perception field. The original VGG16 structure is shown in figure 4.

In 2014, the paper[32] points out that the shallow layers of neural networks mainly extract simple common features of images, such as edge information, and the parameters of these layers do not differ significantly among different data sets after training. While the deep layers mainly extract the characteristics of specific areas, which very much in different tasks. In order to choose which part of vgg16 we should use, we put an image into pretrained vgg16 and visualize part of the feature maps when the image passes through the first, second, and third pool layers. Figure 5 shows the result. From this, we can obtain that layers before the second pool layer only extract very simple information of the images due to the factor that we can recognize the feature maps in figure 3(a) and figure 3(b) that these layers only extract the simple features of the image. But we find the feature maps in figure 3(c) already have extracted some more advanced features of images. Therefore, the layers before the third pool layer of pretrained vgg16 were chosen as our feature extractor. Therefore, our feature extractor contains two two convolution layers plus one pool layer structure, followed by three convolution operations and one pooling operation blocks, as seen in figure 4.

Figure 6 shows the gradient parameters of the first convolution layer and the last convolution layer of the feature extractor and classifier of the model without transfer learning in the process of backpropagation in the training process. From this, we can see that the gradient of the parameters in the classifier is mostly distributed between 0 and 0.02, while the gradient value of the last convolution layer of the feature extractor is between 0 and 0.002, and the gradient value of the parameters in the first layer is between 0 and 0.0008. From this, we can draw that the value of gradient in the process of backpropagation, is reduced. Therefore, in the process of non-transfer learning, the training of feature extractor is much more difficult than the training of the classifier. And in figure 5, we know that the feature extractor only extracts some common low-level features, so the characteristics extracted by different data sets are similar. Therefore, using transfer learning and fixing the parameters of the feature extractor can effectively reduce the difficulty of training, and improves accuracy, especially in the case of a small training set, where it is not easy to train a good feature extractor from the training set which usually occurs in industry field. And in the small training set, the transfer learning feature extractor is more general than the extractor trained by itself, so it can effectively reduce the network's overfitting of very specific features. In the case of only the small data available, the phenomenon of overfitting is often very serious, so transfer learning in the case of small data scale to improve the effect is more obvious. These are the reasons why we use the transfer learning here to overcome the lack of data to training in this problem.

### 3.4. Classifer

The main goal here is to recognize the defect based on the feature maps extracted by the feature extractor. The value of the feature map extracted by the feature extractor is often very large and has great fluctuations. This has a great disturbance to the training of subsequent classifiers. In order to overcome this difficulty, we first execute batch normalization[12] to the input of the feature maps in the classifier to obtain a more numerically appropriate feature map. Batch normalization can not only greatly improves the training speed, accelerate the convergence process, but also increases the classification effect and reduces the difficulty to adjust the hyperparameters.

Dropout value, which is proposed in this paper[14], is also used here to alleviate the overfitting of the neural network. Experiments show that dropout can improve accuracy by 0.03 to 0.04. And in the last part of the classifier, we use the global pool instead of the full connection layer to obtain the probability of a different defect in the steal surface. The full connection layer often contains a very large number of parameters. A large number of parameters can not only lead to greater demand for training data and strong overfitting situations but also make network training more difficult. For example, in vgg16, the number of parameters of the entire network is about 140M, while the number of parameters of the whole connection layer is about 120M, accounting for 0.85 of all parameters. In this article, we try to reduce the parameters to enable the network to have strong capabilities on smaller data sets, so the full connection layer does not meet the purpose of this article. Therefore, we choose the global pool instead of the full connection layer, which is showed clearly in Figure 3.

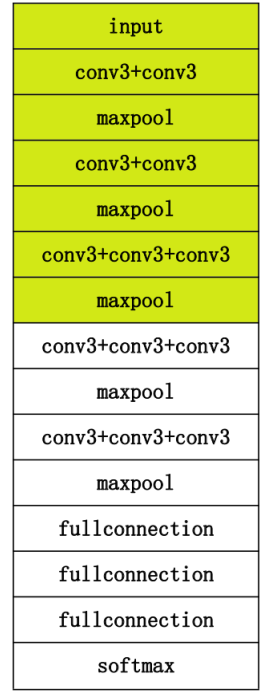

Figure 4. the structure of vgg16(the layers we used here as feature extractor are filled with yellow.)

### 3.5. Test of Our Subsystem

The goal of this research study is to evaluate our method under different conditions using NEU dataset which consists of six different typical surface defects of the hot-rolled steel strip(rolled-in scale (RS), patches(Pa), crazing (Cr), pitted surface (PS), inclusion (In) and scratches(Sc)), and each class has 300 images. We divided this dataset into

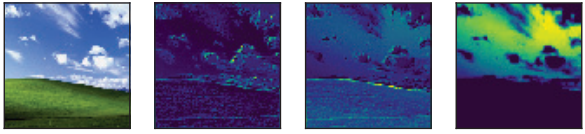

(a) visualization of the feature maps when the image pass through the first pool layer

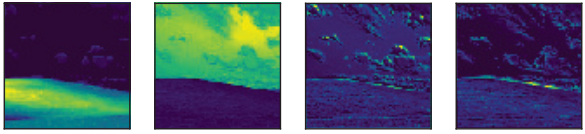

(b) visualization of the feature maps when the image pass through the second pool layer

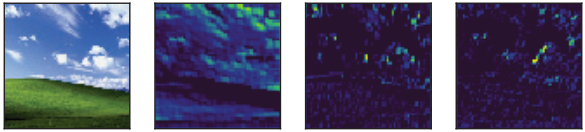

(c) visualization of the feature maps when the image pass through the third pool layer

Figure 5. visualization of part of the feature maps in pretrained vgg16

two equal subsets(training set and test set) and each set has 150 images of each class randomly. In the experiment, we only train the classifier and keep the weights of feature extractor trained by imagenet data set still. We conduct our experiment followed a standard process:

- We preprocess the raw images. First we abstract each channel of the images with its mean. Then we resize the shape of images from $200 \times 200$ into $224 \times 224$ in order fit the input size of the VGG16 neural network.

- We run the all the dataset through the feature extractor to obtrain the feature map of each image.

- Train the classifer using the feature maps of training set and then we use the trained classifer to predict the labels of the feature maps of the test set.

- Compared the predicted labels with ture labels and calculate the presicion of the prediction.

First, we initialize all the parameters in the classifier. All convolution kernels are initialized to a Gaussian distribution with a standard deviation of 0.01 and a mean of 0 and all the biases are initialized to the constant value 0.01. The learning rate is set to a function with an initial value of 0.02 and exponentially decreasing with the training epoch. Each time the parameters are updated 500 times, the learning rate drops to 0.9 of its original value. The initial large learning rate accelerates the learning and reducts time, and the later smaller learning rate contributes to the final convergence. At the same time, we set the save probability of the first dropout in the network to 0.6, and the second to 0.8.

Each time we put three images into the network as a mini-batch. The same as many other classification methods, we use the crossentropy as the loss function. Adam is used to update the weight.

After being trained as above shown, the classifier achieves good performance on the test set. The final accuracy is 99.1 %.

In table 1, we evaluate our method with different scales of the training set, and compare these results with the same neural network without transfer learning to find the effect of transfer learning.In a neural network without transfer learning, we randomly initialize the weight of feature extractor and we train these weights in the training time. As for the training set, we randomly choose n(n=10,30,50,...,150) images of each class in the previous training set and we use the same test set as before.

From table 1 we can obtain that in both cases, when n is less than 70, the accuracy increases rapidly with the increase of n, and the accuracy tends to stabilize when n is greater than 70. It is obvious that this algorithm can achieve good accuracy in smaller data sets, for example, 91.7% accuracy is obtained using only 20 images in each class. At the same time, we found that transfer learning has a more significant improvement in accuracy when the data is smaller.

Table 1. The accuracy of neural network in different scale of dataset with tranfer learning and without transfer learning

| n | 10 | 30 | 50 | 70 | 90 | 110 | 130 | 150 |
|---|---|---|---|---|---|---|---|---|
| accuracy(with transfer learning) | 0.667 | 0.917 | 0.948 | 0.984 | 0.986 | 0.987 | 0.990 | 0.991 |
| accuracy(without transfer learing) | 0.300 | 0.663 | 0.678 | 0.763 | 0.791 | 0.780 | 0.833 | 0.853 |

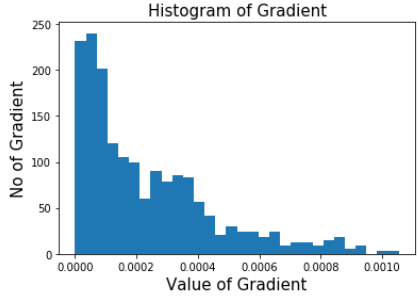
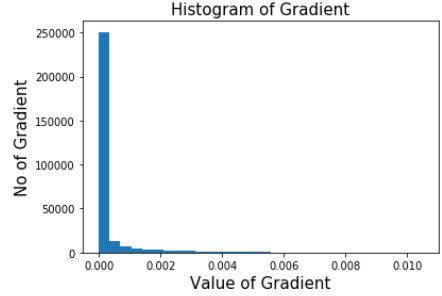

(a) Gradient distribution of the parameters of the first convolution of the feature extractor

(b) Gradient distribution of the parameters of the last convolution of the feature extractor

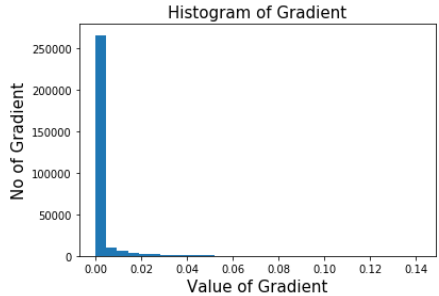
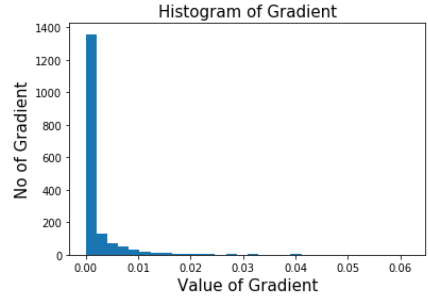

(c) Gradient distribution of the parameters of the classifier's first convolution

(d) Gradient distribution of the parameters of the classifier's last convolution

Figure 6. histogram of gradient for the parameters during the backprogation of non-transfer learning

Table 2. Comparison of defect detection results on test set images using various models(part of the data comes from paper[28]).

| Feature extract Method | Classifier | Accuracy |
|---|---|---|
| LBP | NNC | $95.07 \pm 0.71$ |
|  | SVM | $97.93 \pm 0.66$ |
| LTP | NNC | $95.93 \pm 0.39$ |
|  | SVM | $98.22 \pm 0.52$ |
| CLBP | NNC | $96.91 \pm 0.24$ |
|  | SVM | $98.28 \pm 0.51$ |
| AECLBP | NNC | $97.93 \pm 0.21$ |
|  | SVM | $98.93 \pm 0.63$ |
| CSDC |  | $84.55 \pm 0.05$ |
| PLCNN |  | $87.22 \pm 0.14$ |
| ours |  | $99.10 \pm 0.06$ |

In table 2, we compared our accuracy with the accuracy of the methods mentioned in the pape[28] [35] [5].

As we can seem from the table 2, our method outperforms the traditional methods which manually extract the feature and use the tradition machine learning method to classify these images. Compared with other deep learning methods (CSDC and PLCNN), our method also has higher accuracy. What's more, our accuracy is also more stable than traditional methods, which means that the accuracy doesn't vary too much in each training.

## 4. Improvement in Small Dataset Scale Scene

In section2, our method achieves higher than other methods when each class in the dataset only contains 150 images. But it is not enough, because in practice it is usually difficult to obtain many labeled high-quality data. Therefore, we want to achieve higher accuracy when there are only an extremely small data available. Here we use the case where each class in the dataset only contains 10 images, and we will make a deeper optimization of the model in the following.

## 4.1. Improve the accracy in concern of data augmentation and initialize method

As we all know, the neural network can only find local optimization instead of global optimization; therefore, we will try to use different data augmentation and initialization methods to make the optimizer of the neural network find better optimization. More specifically, we will consider the brightness of the images, the position of the defect in the images and initialize methods we use to initialize the classifier here.

When there are only a few images, the images of one class may be all dim or bright. Therefore, the neural network may take the brightness of the picture as a feature in the process of learning. This is not in line with the facts. We know that the brightness of the picture is represented by the value of the picture pixel. In order to prevent this situation, we multiply all pixels by 1.2, 1.4, 1.6 plus 10, and assign the value of 255 to pixels who are greater than 255 after processing. After processing, we put images with different brightness together. we train all these images together. The final result was 0.719, which was 0.052 higher than the original accuracy.

In addition to brightness, defects are also an important feature in the location of the picture. A neural network can easily fit this positional feature during learning, especially when the number of pictures is very small, in which often a type of defect may be located in the same position of images. Thus the neural network mistakenly believes that the positions of defects are important factors to classify. To overcome this overfitting, we flip all images left and right, vertically and left and right, and train the resulting image with the original image. Using this method, our accuracy was increased from 0.667 to 0.805. What's more, we also rotate this image with a different angle, which increases the accuracy to 0.824. As shown in table .

Combined with all these methods above, we finally obtained 0.867 accuracy, which is 0.2 higher than the original. This shows that data augmentation can help improve the accuracy greatly when the data set is small. And compare these methods, we can find that the neural network is easier to be misled by the position of the defect than the brightness of the image.

Here, four initialize methods(Gaussian distribute with 0 mean and 0.01 std, uniform distribution for -0.01 to 0.01, Xavier[6] and MSRA[11]) are used to find the influence of initializing methods on the accuracy. All these methods are commonly used today. Gaussian distribution and uniform distribution are the simplest methods of initialization.

Because the neural network has more parameters, in order to prevent the initial output value after too large, making the initial training difficult, we assign the parameters to a smaller value, that's why we choose Gaussian distribution with 0 mean and 0.01 std and uniform distribution for -0.01 to 0.01. The Xavier initialization method was proposed by Xavier Glorot et all in order to prevent activation function saturating, which is very bad to prevent the backpropagation of gradients.MSRA initialization is a refined version of xavier method, which makes it much better in the RELU network.

Table 4 shows the result of using different initialize methods. All of the input images are preprocess using data augmentation. As we can see, when the dataset is small, initialized methods have a strong influence on accuracy. Xavier performs best here with an accuracy of 0.921.

In general, we can know from this chapter that when choosing the data augmentation methods combining all four traditional methods and the initialize method named Xavier, we can get the best result for our algorithm, whose training data set is only ten images.

## 4.2. Analyze the effect of noise

Here, we try to analyze the effect of noise on the accuracy. In order to do this, we add different noise (5dB,30dB) to the images manually and obtain the accuracies in different scale of the training set(all the images used here are already be preprocessed by data augmentation and we use the best initialize method here).

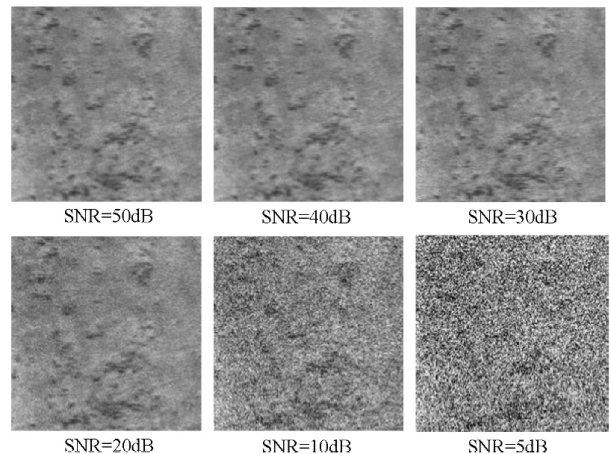

Figure 7. Surface defect images polluted by Gaussian noise with different SNR.

The result is shown in Figure 8.

From Figure 8, we can know that the small noise(30dB)

Table 3. The accuracy of neural network after using different data augmentation(10 images each class)

| method | original | brightness change | image flip | image rotate | combined all |
|---|---|---|---|---|---|
| accuracy | 0.667 | 0.719 | 0.805 | 0.824 | 0.867 |

Table 4. The accuracy of neural network using different initialize(10 images each class) methods

| method | Guassian distribution(0,0.01) | uniform distribution | Xavier | MSRA |
|---|---|---|---|---|
| accuracy | 0.867 | 0.896 | 0.921 | 0.912 |

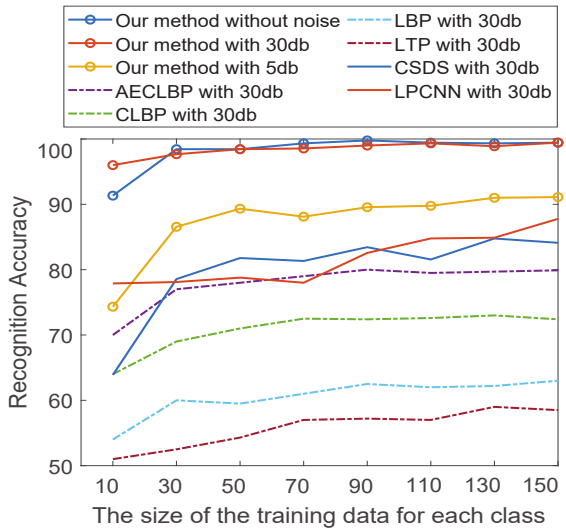

Figure 8. The recognition accuracy with different number of the training test for each class under different noise using different methods(part of the data comes from [28])

## 5. Discussion

### 5.1. Advantage

- This model takes the trained VGG16 for feature extraction, effectively reducing the demand for data sets. we achieve the accuracy of 0.96 when each class contains only 10 images, which makes it suitable for use in small or specialized plants where high-quality images are difficult to obtain.

- The construction of this method does not require the expertise of a specific field, so it is convenient for the model to use in different fields.

- Our method is light, which means that it does not contain too many parameters and does not need too many resources to train. As a result, it will be easy to be taken into practice.

can improve the accuracy when each class contains 10 images due to the factor that small noise can add some uncertainty which can alleviate the over-fitting. Nevertheless, too much noise can make the model hard to fit the training set and lower the accuracy. But the noise's effect is not very much. In the small noise, we obtain our highest accuracy with the scale of 10 images per class under 30dB noise.

Compared with previous methods in [28], our method has a strong resistance to the noise. The accuracy of the previous drop very much even if we add a small noise, as shown in figure 5 that only 30db noise can cause a strong decrease in accuracy. When we add noise, the classifier can try to adapt the noise in the training process. Because when we add noise, our method, as well as other deep learning methods, can get used to this noise in the training process.

Using the data augmentation and arxiv initialization methods, we obtain our best accuracy (96.0 %) in the dataset where each class only has 10 images when we add small noise(30db) to the training images.

### 5.2. Disadvantage and Future work

The original picture in this model is a 200 x 200x 1 grayscale picture, but because the vgg16 model requires an input of 224 x 224 x 3 picture size. So we repeat the picture three times so that each channel of the picture contains the same information. This method makes the same information repeated and reduces the efficiency of the model. In the future, it may be considered that replaces the second and third channels of the image by feature map of this image, such as the feature map after the canny edge detection of the images. However, this change will make the dataset we used extremely different from the image we use to pretrained vgg16, so it is necessary to make appropriate fine-tuning of vgg16.

Using the data augmentation and Xavier initialization methods, we obtain our best accuracy (96.0 %) in the dataset where each class only has 10 images when we add small noise(30db) to the training images.

# References

[1] Anzar Alam, Jan Thim, Mattias ONils, Anatoliy Manuilskiy, Johan Lindgren, and Joar Lidén. Online surface characterization of paper and paperboards in a wide-range of the spatial wavelength spectrum. *Applied Surface Science*, 258(20):7928–7935, 2012.

[2] Francisco G Bulnes, Rubén Usamentiaga, Daniel F García, and Julio Molleda. Vision-based sensor for early detection of periodical defects in web materials. *Sensors*, 12(8):10788–10809, 2012.

[3] Shin-Min Chao and Du-Ming Tsai. An anisotropic diffusion-based defect detection for low-contrast glass substrates. *Image and Vision Computing*, 26(2):187–200, 2008.

[4] Theodore V Galambos and Mayasandra K Ravindra. Properties of steel for use in lrfd. *Journal of the Structural Division*, 104(9):1459–1468, 1978.

[5] Yiping Gao, Liang Gao, Xinyu Li, and Xuguo Yan. A semi-supervised convolutional neural network-based method for steel surface defect recognition. *Robotics and Computer-Integrated Manufacturing*, 61:101825, 2020.

[6] Xavier Glorot and Yoshua Bengio. Understanding the difficulty of training deep feedforward neural networks. In *Proceedings of the thirteenth international conference on artificial intelligence and statistics*, pages 249–256, 2010.

[7] Kasthurirangan Gopalakrishnan, Siddhartha K Khaitan, Alok Choudhary, and Ankit Agrawal. Deep convolutional neural networks with transfer learning for computer vision-based data-driven pavement distress detection. *Construction and Building Materials*, 157:322–330, 2017.

[8] Zhenhua Guo, Lei Zhang, and David Zhang. A completed modeling of local binary pattern operator for texture classification. *IEEE transactions on image processing*, 19(6):1657–1663, 2010.

[9] Kun He and Li Wang. A review of energy use and energy-efficient technologies for the iron and steel industry. *Renewable and Sustainable Energy Reviews*, 70:1022–1039, 2017.

[10] Kaiming He, Xiangyu Zhang, Shaoqing Ren, and Jian Sun. Deep residual learning for image recognition. 2015.

[11] Kaiming He, Xiangyu Zhang, Shaoqing Ren, and Jian Sun. Delving deep into rectifiers: Surpassing human-level performance on imagenet classification. In *Proceedings of the IEEE international conference on computer vision*, pages 1026–1034, 2015.

[12] Sergey Ioffe and Christian Szegedy. Batch normalization: Accelerating deep network training by reducing internal covariate shift. *arXiv preprint arXiv:1502.03167*, 2015.

[13] Yi Jiang, Tung-Chai Ling, Caijun Shi, and Shu-Yuan Pan. Characteristics of steel slags and their use in cement and concretela review. *Resources, Conservation and Recycling*, 136:187–197, 2018.

[14] Alex Krizhevsky, Ilya Sutskever, and Geoffrey E. Hinton. Imagenet classification with deep convolutional neural networks. In *International Conference on Neural Information Processing Systems*, 2012.

[15] Fang Li, Stefano Spagnul, Victor Odedo, and Manuchehr Soleimani. Monitoring surface defects deformations and displacements in hot steel using magnetic induction tomography. *Sensors*, 19(13):3005, 2019.

[16] Wu-bin Li, Chang-hou Lu, and Jian-chuan Zhang. A local annular contrast based real-time inspection algorithm for steel bar surface defects. *Applied Surface Science*, 258(16):6080–6086, 2012.

[17] Xiaoli Li, Shiu Kit Tso, Xin-Ping Guan, and Qian Huang. Improving automatic detection of defects in castings by applying wavelet technique. *IEEE Transactions on Industrial Electronics*, 53(6):1927–1934, 2006.

[18] Zheng Liu, Xusheng Wang, and Xiong Chen. Inception dual network for steel strip defect detection. In *2019 IEEE 16th International Conference on Networking, Sensing and Control (ICNSC)*, pages 409–414. IEEE, 2019.

[19] Hiroshi Maki, Y Tsunozaki, and Y Matsufuji. Magnetic online defect inspection system for strip steel. *Iron and Steel Engineer(USA)*, 70(1):56–59, 1993.

[20] Robert C Makkus, Arno HH Janssen, Frank A de Bruijn, and Ronald KAM Mallant. Use of stainless steel for cost competitive bipolar plates in the spfc. *Journal of power sources*, 86(1-2):274–282, 2000.

[21] David Martin, Domingo Miguel Guinea, María C García-Alegre, E Villanueva, and Domingo Guinea. Multi-modal defect detection of residual oxide scale on a cold stainless steel strip. *Machine Vision and Applications*, 21(5):653–666, 2010.

[22] R Narayanan and IYS Darwish. Use of steel fibers as shear reinforcement. *Structural Journal*, 84(3):216–227, 1987.

[23] Timo Ojala, Matti Pietikäinen, and Topi Mäenpää. Multiresolution gray-scale and rotation invariant texture classification with local binary patterns. *IEEE Transactions on Pattern Analysis & Machine Intelligence*, (7):971–987, 2002.

[24] Sinno Jialin Pan and Qiang Yang. A survey on transfer learning. *IEEE Transactions on Knowledge Data Engineering*, 22(10):1345–1359, 2009.

[25] G Rosati, G Boschetti, A Biondi, and A Rossi. Real-time defect detection on highly reflective curved surfaces. *Optics and Lasers in Engineering*, 47(3-4):379–384, 2009.

[26] Luke Scime and Jack Beuth. A multi-scale convolutional neural network for autonomous anomaly detection and classification in a laser powder bed fusion additive manufacturing process. *Additive Manufacturing*, 24:273–286, 2018.

[27] Karen Simonyan and Andrew Zisserman. Very deep convolutional networks for large-scale image recognition. *Computer Science*, 2014.

[28] Kechen Song and Yunhui Yan. A noise robust method based on completed local binary patterns for hot-rolled steel strip surface defects. *Applied Surface Science*, 285(21):858–864, 2013.

[29] Christian Szegedy, Wei Liu, Yangqing Jia, Pierre Sermanet, Scott Reed, Dragomir Anguelov, Dumitru Erhan, Vincent Vanhoucke, and Andrew Rabinovich. Going deeper with convolutions. 2014.

[30] Xiaoyang Tan and Bill Triggs. Enhanced local texture feature sets for face recognition under difficult lighting conditions. In *International workshop on analysis and modeling of faces and gestures*, pages 168–182. Springer, 2007.

[31] Lei Wang, Ke Xu, and Peng Zhou. Online detection technique of 3d defects for steel strips based on photometric stereo. In *2016 Eighth International Conference on Measuring Technology and Mechatronics Automation (ICMTMA)*, pages 428–432. IEEE, 2016.

[32] Jason Yosinski, Jeff Clune, Yoshua Bengio, and Hod Lipson. How transferable are features in deep neural networks? *Eprint Arxiv*, 27:3320–3328, 2014.

[33] Jong Pil Yun, SungHoo Choi, Jong-Wook Kim, and Sang Woo Kim. Automatic detection of cracks in raw steel block using gabor filter optimized by univariate dynamic encoding algorithm for searches (udeas). *NDT & E International*, 42(5):389–397, 2009.

[34] Jong Pil Yun, SungHoo Choi, and Sang Woo Kim. Vision-based defect detection of scale-covered steel billet surfaces. *Optical Engineering*, 48(3):037205, 2009.

[35] Shiyang Zhou, Youping Chen, Dailin Zhang, Jingming Xie, and Yunfei Zhou. Classification of surface defects on steel sheet using convolutional neural networks. *Mater. Technol*, 51(1):123–131, 2017.